\documentclass[letterpaper, 10 pt, conference]{ieeeconf}  

\IEEEoverridecommandlockouts                              

\usepackage{graphics} 
\usepackage{epsfig} 
\usepackage{mathptmx} 
\usepackage{times} 
\usepackage{amsmath} 
\usepackage{amssymb}  
\usepackage{subcaption}
\usepackage{multirow}
\usepackage{algorithm}

\title{\LARGE \bf
Inside-out Infrared Marker Tracking via Head Mounted Displays for Smart Robot Programming}

\author{David Puljiz$^{1}$, Alexandru-George Vasilache$^{1}$, Michael Mende$^{1}$, Bj\"orn Hein$^{1,2}$
\thanks{$^{1}$Intelligent Process Automation and Robotics Lab (IPR), Institute for Anthropomatics and Robotics, Karlsruhe Institute of Technology, Karlsruhe, Germany
        {\tt\small david.puljiz@kit.edu}}%
\thanks{$^{2}$ Karlsruhe University of Applied Sciences, Karlsruhe, Germany}
}

\begin{document}

\maketitle
\thispagestyle{empty}
\pagestyle{empty}

\begin{abstract}

Intuitive robot programming through use of tracked smart input devices relies on fixed, external tracking systems, most often employing infra-red markers. Such an approach is frequently combined with projector-based augmented reality for better visualisation and interface. The combined system, although providing an intuitive programming platform with short cycle times even for inexperienced users, is immobile, expensive and requires extensive calibration. When faced with a changing environment and large number of robots it becomes sorely impractical. Here we present our work on infra-red marker tracking using the Microsoft HoloLens head-mounted display. The HoloLens can map the environment, register the robot on-line, and track smart devices equipped with infra-red markers in the robot coordinate system. We envision our work to provide the basis to transfer many of the paradigms developed over the years for systems requiring a projector and a tracked input device into a highly-portable system that does not require any calibration or special set-up. We test the quality of the marker-tracking in an industrial robot cell and compare our tracking with a ground truth obtained via an ART-3 tracking system.   

\end{abstract}

\section{INTRODUCTION}

Fast and intuitive robot programming, even by lay-users, would bring massive changes to the industrial robot sector. Cutting down the cost and time required for programming and setup would mean easier implementation of flexible production paradigms, increased customisation of products and larger acceptance of robotics and automation in small and medium sized enterprises \cite{schraft2006need}. Thus, intuitive robot programming has been a staple of robotics research for years. \par

One particular research direction is the use of varied intuitive input devices to facilitate robot programming. These devices are tracked by external tracking systems, most notably using infra-red (IR) markers. \textit{Hein et al.} \cite{hein2009intuitive} have shown that such devices allow even untrained people to solve robot jogging, trajectory input and surface interaction error-free. Such devices can also feature additional sensors, such as a force sensor in the tip, to help with input. \par

Tracked input devices have often been coupled with projector-based augmented reality (AR) to provide increased feedback to the user. In \cite{Zaeh2007Interactive} the user could add waypoints, select and edit preexisting waypoints and add geometric features by defining characteristic points. A small user study with 9 male participants of medium robot programming experience has shown a more than five-fold decrease in the programming time of simple trajectories. \par

Reihart \textit{et al.} \cite{Reinhart2007projection-based} discuss the interaction possibilities of such a system. They list trajectory editing, interaction with virtual menus and the digitising of object surfaces - generating 3D models of unknown objects on the fly, as interaction options. \par

\textit{Gaschler et al.} \cite{gaschler2014intuitive} show a system for general trajectory and obstacle input, enhancing the system by providing an additional monitor with a virtual 3D view of the robot arm, it's path and the defined obstacles. \par

\textit{Reinhart et al.} \cite{REINHART2008programing} as well as \textit{Leutert et al.} \cite{Leutert2013} apply such a system for welding applications where the path was planned on obstacle free surfaces. In particular \textit{Reinhart et al.} show a 30\% increase in cycle times when using their system. \par

\begin{figure}[t]

    \includegraphics[width=0.48\textwidth]{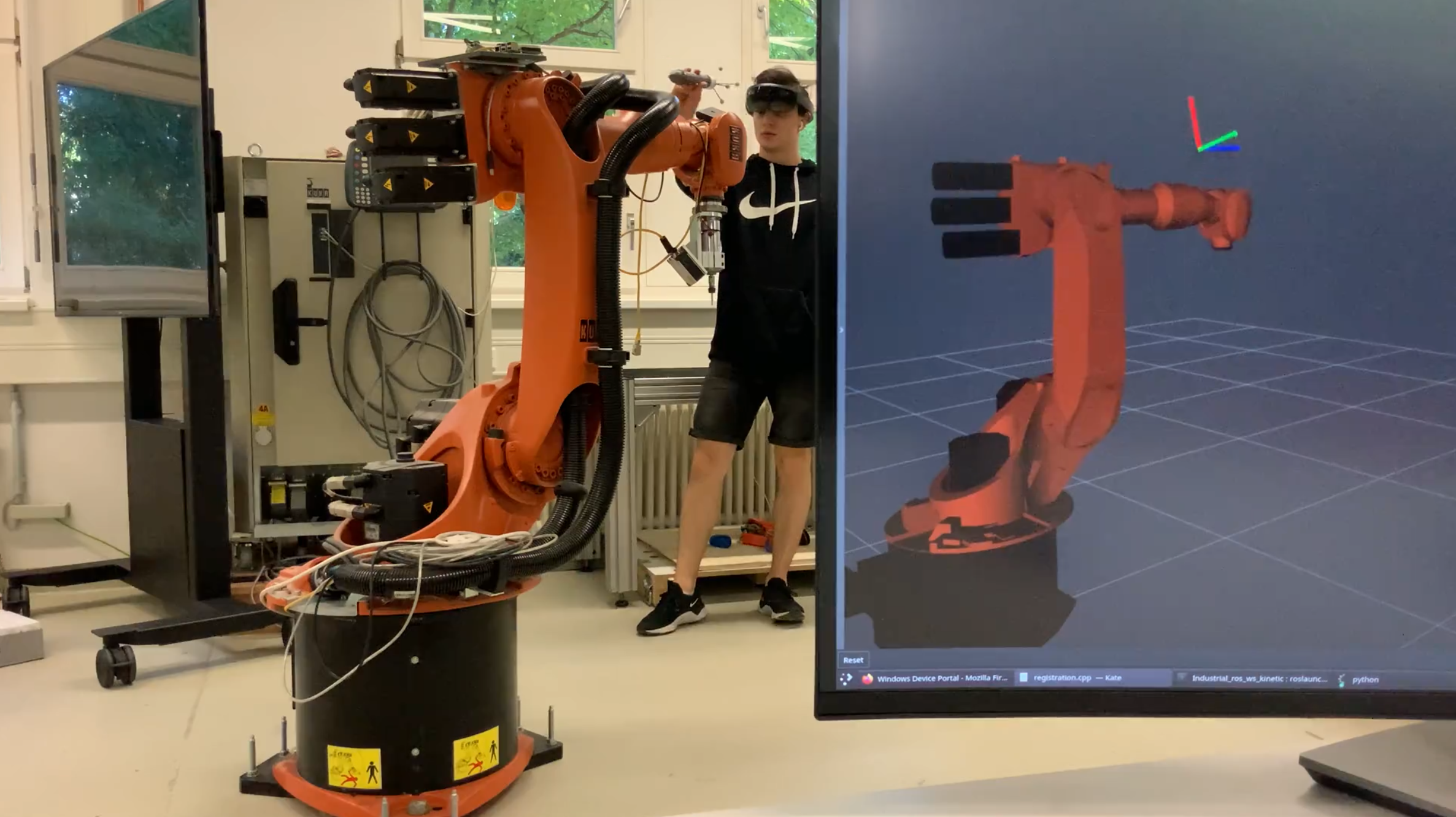}
    \caption[]{The user wearing the HoloLens and holding the tracked input device equipped with IR-markers. The input device is tracked purely through the on board sensors of the HoloLens. The robot is referenced at the start of the interaction using a quick, on-line interactive referencing step. One can also see the live tracking in RViz on the screen of a desktop PC running ROS.}
    \label{fig:intro}
\end{figure}

In \cite{notheis2012ar-based} a head-mounted display (HMD) was used instead of a projector. The system was used to program a trajectory of an industrial robot carrying a suspended load, in this case a ball. Once a trajectory input is given, a model-based control algorithm is tested in simulation and visualised in AR so the user can check the stability of the input trajectory. \par 

As can be seen, systems with projector-based AR coupled with tracked input devices have been shown to increase intuitiveness and speed of robot programming and reprogramming. Such systems however require extensive calibration of the projector and IR tracking system. This translates to long set-up times and costs making them unfeasible in flexible and changing production paradigms. Furthermore such systems are immobile and thus become impractical and even more costly when large number of robots are involved. \par

The system proposed in this paper, on the other hand, is fully portable, robot-agnostic and does not require any calibration procedure. The programmer can freely take the system from robot cell to robot cell and program individual robots. \par

It consists only of a Microsoft HoloLens HMD, an input device equipped with IR markers, in our case a pen, and a desktop computer connected to the robot and running the robot operating system (ROS). The HoloLens can freely move around and track the input device. All the tracking and computation are done on the device itself. \par 

The HoloLens can interface with different computers to acquire the specific robot description, model and joint state. This information is then used to reference the robot on-line by registering a generated model point cloud to an environmental point cloud obtained through the use of the HoloLens' depth sensor. The reference allows complete transformation between the coordinate systems of the HoloLens and the robot. This allows the HoloLens to send back its pose and the pose of the tracked input device in the robot's coordinate system. \par

Our work here forms a basis to transfer all of the previously developed programming and interaction modalities into a fully portable and flexible system, negating the previous downsides of such systems.\par

\begin{figure}[h]

    \includegraphics[width=0.48\textwidth]{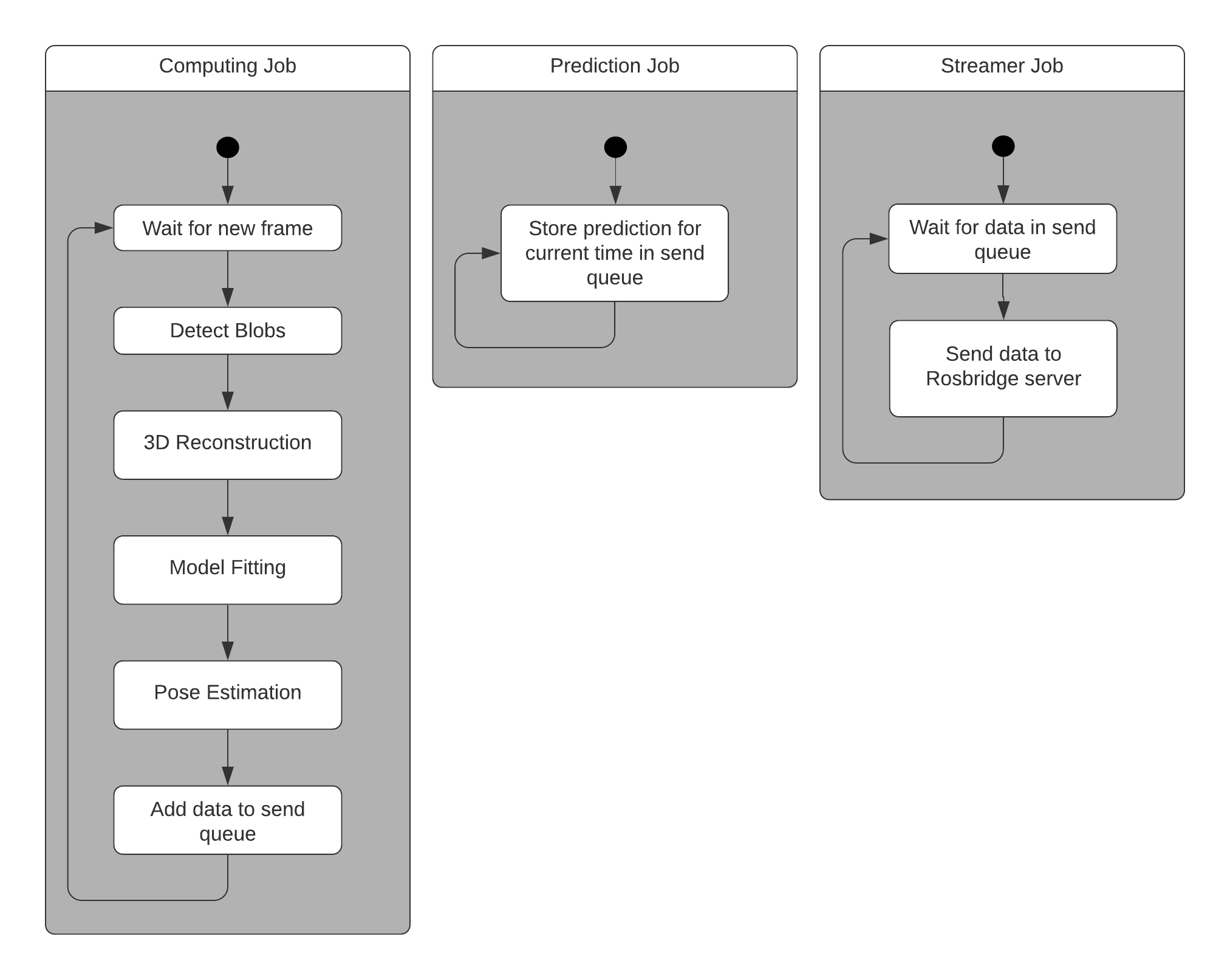}
    \caption[]{The workflow of the marker tracking algorithm is split into three jobs or processes - The Computing Job in charge of detecting markers in reflectivity frames and computing the input device's pose. The Prediction Job that interpolates the object's position and orientation in between frames. And the streamer job that streams the tracking data to the desktop computer running ROS. The Computing Job being computationally expensive runs on two threads, while other jobs run each on one.}
    \label{fig:workflow}
\end{figure}

\section{METHODOLOGY}

\subsection{Mapping}

The HoloLens provides two modes of depth sensing, the short-throw, with 30 frames per second update rate and a range of 0.2-1 meters, and the long-throw, with one to five frames per second update rate and a range of 0.5-4 meters. These modes work in parallel and each one provides a reflectivity and a depth stream. The reflectivity stream is the raw data of how much IR light was reflected back into the sensor after the environment was illuminated by IR emitters. The depth stream is the filtered and processed stream where each pixel corresponds to the distance form the camera. \par

For mapping we use the long throw depth stream, as this eliminates any interference from the tracked device and the user's hands. The point cloud is then generated by taking the location of the HoloLens when the frame was captured and projecting the values of the depth frame using a pinhole camera model. Registration between different frames was found not to be necessary. The resulting point cloud is down-sampled using voxel-grid filtering to ensure uniform point density. It is then filtered with an outlier removal filter, removing any point that had less than 9 neighbours in a radius of 5 cm. The point cloud was then smoothed with moving least squares \cite{Alexa2003MLS}. RANSAC plane detection was used to detect planes and map all the points near the plane to the plane itself. This improved the resolution of objects near flat surfaces. \par

This point cloud is then used in referencing to register the robot model point cloud. It can also be used to create an octomap representation of the robot environment and thus prevent any possible collisions with the environment during programming. More detail on the mapping, the experiments and the octomap generation can be found in \cite{puljiz2020hololens}. \par 

\subsection{Referencing}

When the HoloLens connects to the desktop computer running ROS, the universal robot description file (urdf), the robot model meshes and the joint states are used to create the current model point cloud of the robot. The user is asked to place a seed hologram near the robot's base as the first guess to the registration algorithm. This is done to prevent the registration algorithm getting stuck in local minima as it tends to happen with algorithms based on local features. It was found that the simple Iterative Closest Point (ICP) algorithm performs very well and at high speeds. Once the registration step is done the transformation between the HoloLens world coordinate frame and the robot's world coordinate frame is calculated. This allows the transformation of the HoloLens' pose and thus also of the tracked input device into the robot's coordinate system. \par  

\begin{figure*}[h]
\centering
\vspace {2pt}
\subcaptionbox{}
[0.43\textwidth]{\includegraphics[width=0.43\textwidth]{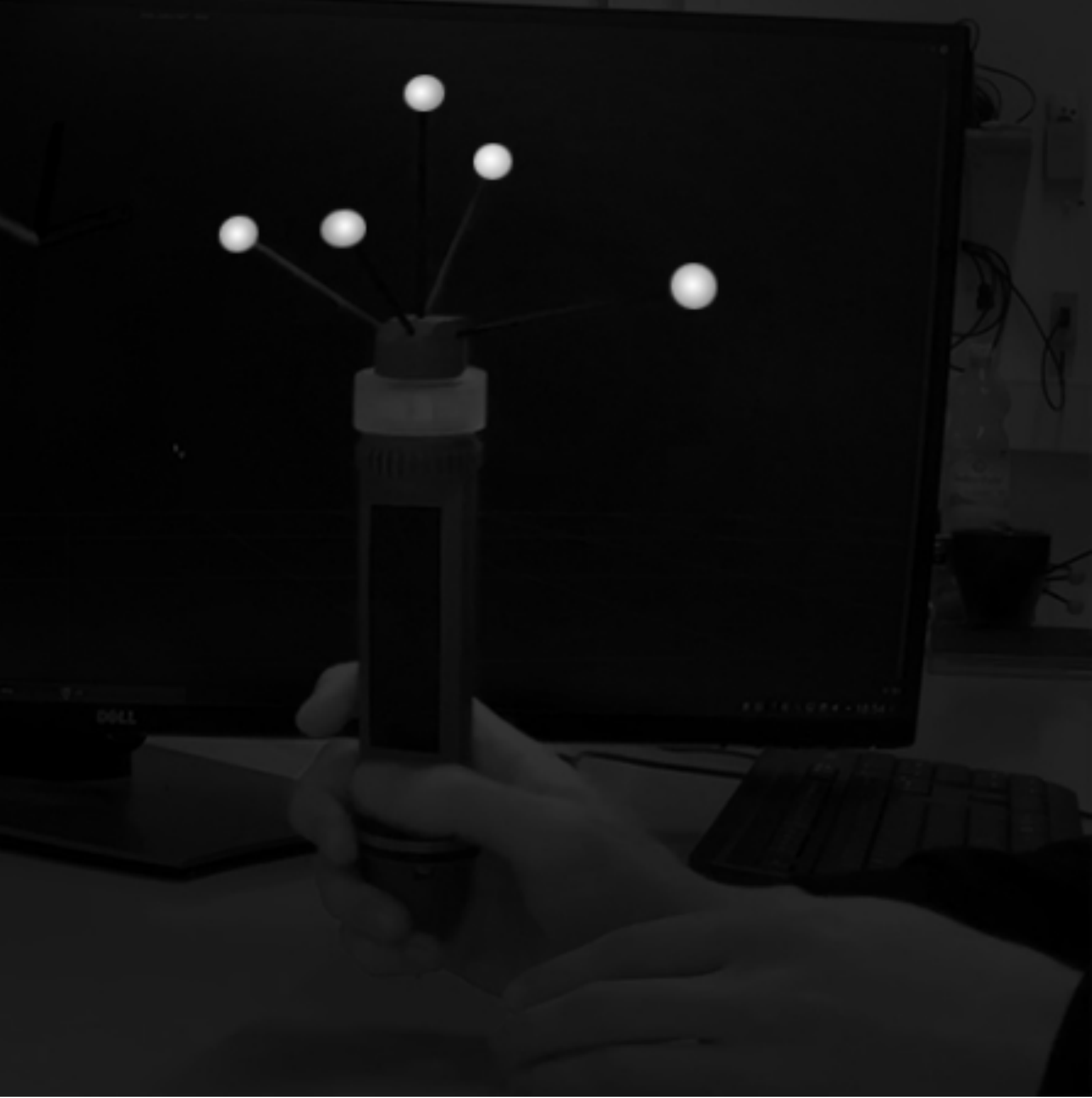}}
\subcaptionbox{}
[0.43\textwidth]{\includegraphics[width=0.43\textwidth]{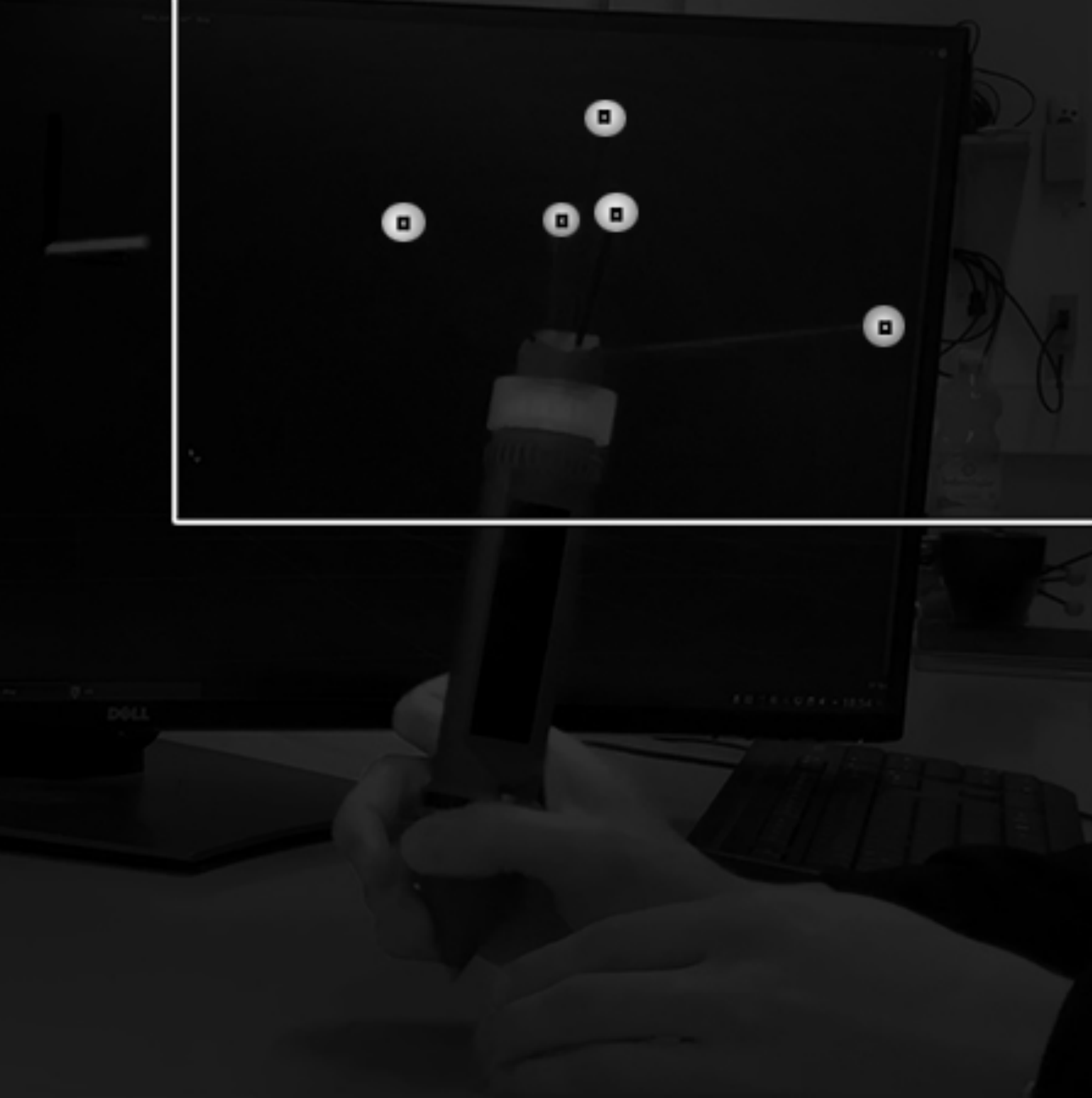}}
    \caption[]{(a) The original, distorted reflectivity stream. One can notice the IR markers illuminated by the HoloLens' depth senor are quite visible. (b) The detected position of the blobs and the defined region of interest for the next frame.}
    \label{fig:tracking}
\end{figure*}  

\subsection{Tracking}

For marker tracking we use the short-throw reflectivity stream to track IR markers illuminated by the HoloLens' depth sensor. \par

After undistorting the frame \cite{TUW-210294} we apply a threshold filter, as the markers are highly reflective. The threshold filter is applied so that only pixel values higher than 250 remain, since the marker pixel values are very close to 255. This will eliminate interference from other reflective surfaces. The image size is reduced by half for blob detection to save on processing speed. The centre pixels ($x_i$,$y_i$) of each blob are then mapped into 3D space using the pinhole camera model: \par

\begin{equation}
 m_i = \left( \frac{Z_i \cdot x_i}{f}, \, \frac{Z_i \cdot y_i}{f}, \, Z_i \right) ^T
\end{equation}

Where f is the focal length of the HoloLens depth camera, and $Z_i$ is the distance of the i-th blob centre from the camera, obtained by mapping the centre pixel of the blob to the depth stream of the same frame. \par

The distance matrix ($D_R$) of the set of detected markers $M_R$ is created and the distances compared to the distance matrix of the model ($D_M$). Each distance matrix element $d^{ij}$ holds the distance between the points $m^i$ and $m^j$. As there will inevitably be a reconstruction error, a tolerance threshold $\delta$ is introduced. Two distances are considered equal if: \par

\begin{equation} \label{equ:distance_threshold}
\mid d^{pq}_R - d^{ij}_M \mid \, < \delta
\end{equation}

Where $d^{pq}_R$ is the distance between two reconstructed markers, $m^p_R$ and $m^q_R$ and $d^{ij}_M$ is the distance between two model points, $m^i_M$ and $m^j_M$. \par

After each of the found markers has been assigned to a correspondent in the model, the rotation and translation between the two rigid bodies has to be determined. \par 

The main approaches are through the use of orthonormal rotational matrices \cite{article} and the singular value decomposition (SVD) of the covariance matrix \cite{4767965}. \textit{Umeyama} (\cite{Umeyama1991LeastSquaresEO}, \textit{Kanatani} \cite{291441} and \textit{Challis} \cite{c0ae9ca782784f238570e1fc0c232744}, improved the previous methods, especially by fixing a flaw where noisy data sometimes caused the rotation matrix to have a determinant of $-1$ and thus become a reflection rather than a rotation. Our algorithm is based on the work from \textit{Kanatani} \cite{291441} and \textit{Challis} \cite{c0ae9ca782784f238570e1fc0c232744} who both independently expanded on the method proposed by \textit{Arun et al.} in \cite{4767965}. \par

Given a point $x_i$ in the set $M_R$ and point $y_i$ in the set $M_M$ ($i = 1, 2, ..., N$ and $N \geq 3$). The transformation between them is:
\begin{equation}
y_i = R x_i + t
\end{equation}

Where $R$ is a 3x3 rotation matrix and $t$ is the translation vector. \par

The least-squares problem of finding $R$ and $t$ is equivalent to minimising the following expression: \par

\begin{equation} \label{equ:min_probl}
\frac{1}{n} \displaystyle\sum\limits_{i=1}^n (R x_i + t - y_i)^2 = 
\frac{1}{n} \displaystyle\sum\limits_{i=1}^n (R x_i + t - y_i)^T (R x_i + t - y_i)
\end{equation}

This can be further simplified by eliminating $t$ as an unknown variable. We compute the mean vectors of both point sets $\overline{x}$ and $\overline{y}$. The vector $t$ can then be computed as: \par

\begin{equation}
t = \overline{y} - R\overline{x}
\end{equation}
            
Substituting $t$ in Equation \ref{equ:min_probl}, we get: \par
\begin{equation} \label{equ:first_sub}
\frac{1}{n} \displaystyle\sum\limits_{i=1}^n
(R x_i - y_i + \overline{y} - R\overline{x})^T
(R x_i - y_i + \overline{y} - R\overline{x})
\end{equation}
            
We define two new point sets by translating the sets $y_i$ and $x_i$ so that the means $\overline{x}'$ and $\overline{y}'$ of these two new point sets are located in the origins of the two reference frames: \par

\begin{equation}
x'_i = x_i - \overline{x}, \quad y'_i = y_i - \overline{y}
\end{equation}

\begin{figure*}[h!]
\centering
\vspace {2pt}
\subcaptionbox{}
[0.47\textwidth]{\includegraphics[width=0.47\textwidth]{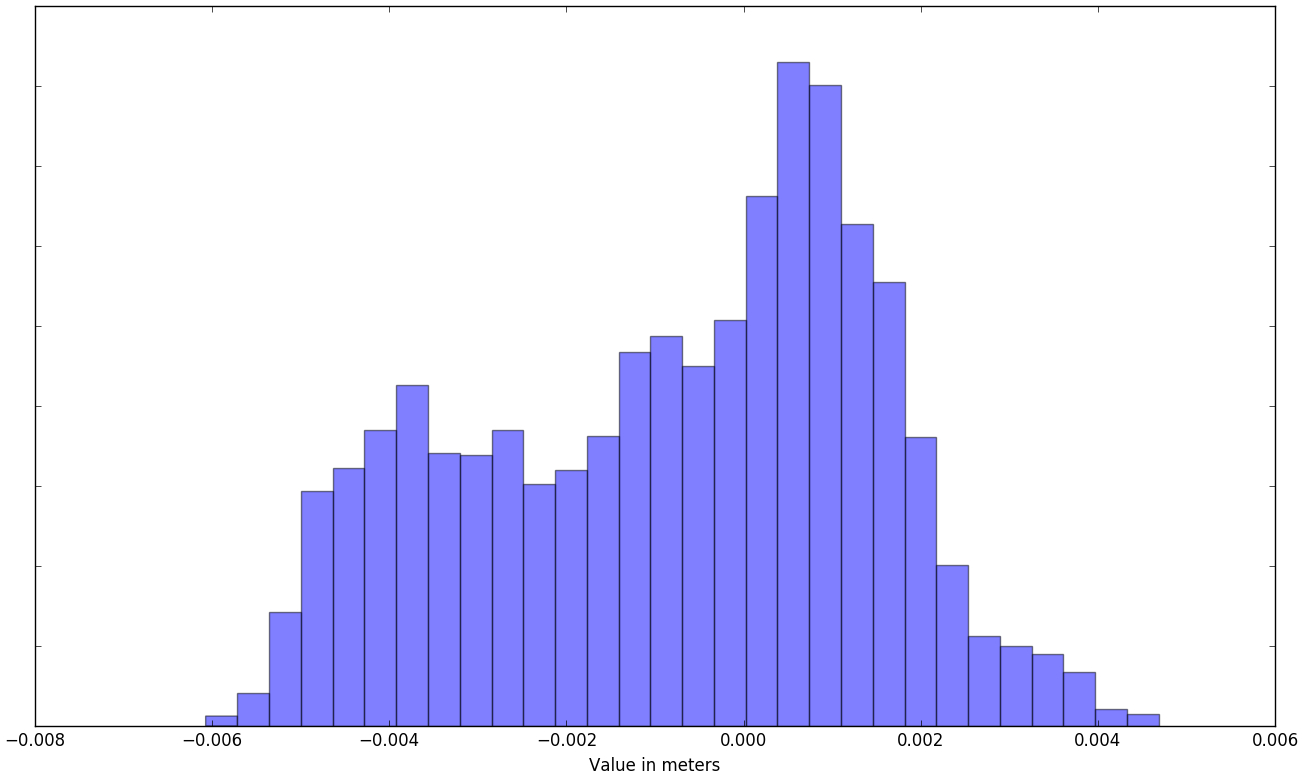}}
\subcaptionbox{}
[0.46\textwidth]{\includegraphics[width=0.46\textwidth]{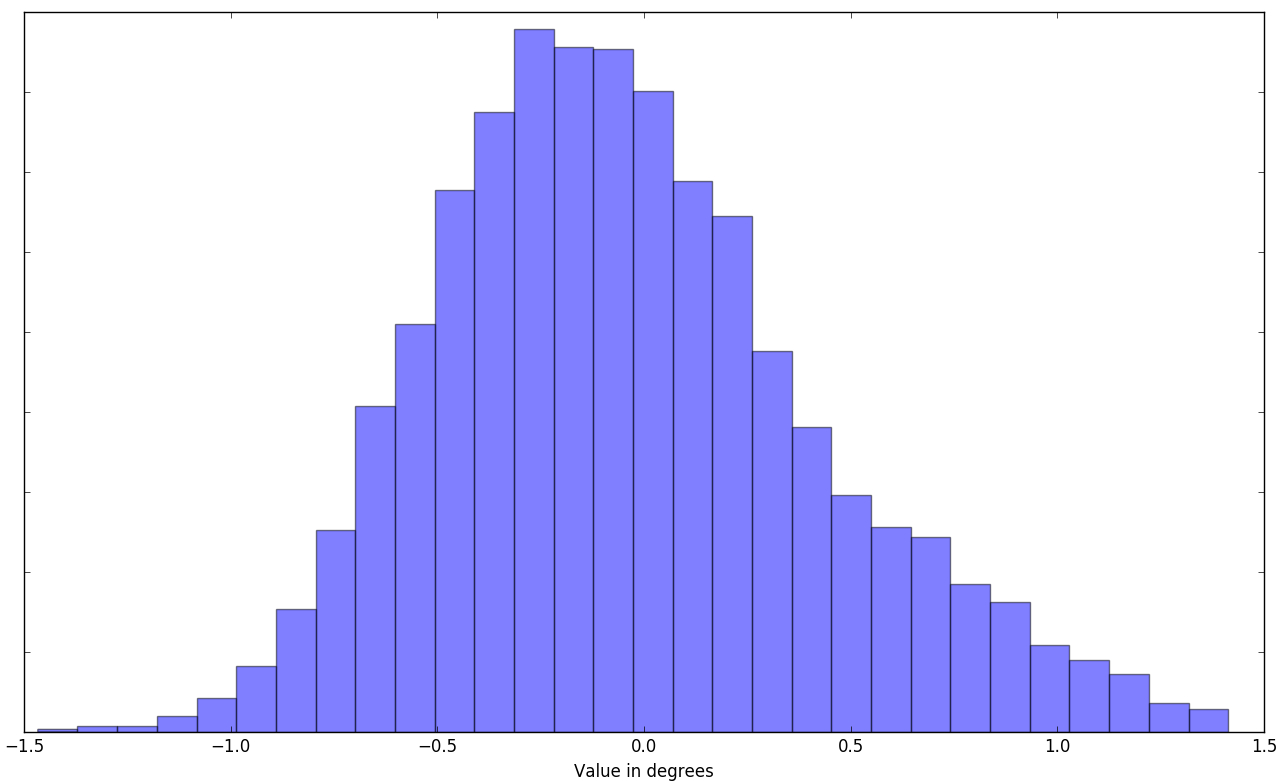}}
    \caption[]{(a) Distribution of the positional error of 5000 data points. The average absolute error is of 1.9mm (b) Distribution of the angular error of 5000 data points. The average absolute error is of $0.37^{\circ}$. To note is the non-Gaussian distribution of error due to the IMU drift being the major component of error.}
    \label{fig:static_err}
\end{figure*}

Substituting these two new vectors in Equation \ref{equ:first_sub}, we get: \par

\begin{equation}
\frac{1}{n} \displaystyle\sum\limits_{i=1}^n
(y'_i - R x'_i)^T
(y'_i - R x'_i)
\end{equation}
            
By expanding and reducing the equation above, we get: \par

\begin{equation}
\frac{1}{n} \displaystyle\sum\limits_{i=1}^n
({y'_i}^T y'_i + {x'_i}^T x'_i - {2y'_i}^T R x'_i )
\end{equation}

By using the following equivalences: \par

\begin{equation}
 \{ R x'_i \}^T y'_i = {y'_i}^T R x'_i
\end{equation}

\begin{equation}
\{ R x'_i \}^T R x'_i = {x'_i}^T R^T R x'_i = {x'_i}^T x'_i
\end{equation}

Thus, minimising Equation \ref{equ:min_probl} is equivalent to maximising: \par

\begin{equation}
 \frac{1}{n} \displaystyle\sum\limits_{i=1}^n
 ({y'_i}^T R x'_i )
\end{equation}
            
Rearranging and summing this, gives the following to maximise: \par

\begin{equation} \label{equ:trace}
\frac{1}{n} \displaystyle\sum\limits_{i=1}^n
({y'_i}^T R x'_i ) = 
tr \left( R^T \frac{1}{n} \displaystyle\sum\limits_{i=1}^n {y'_i} {x'_i}^T \right) = 
tr (R^T C)
\end{equation}

Where $tr( )$ is the trace of a matrix and $C$ is the correlation matrix computed as: \par

\begin{equation}
C = \frac{1}{n} \displaystyle\sum\limits_{i=1}^n (y_i - \overline{y})
(x_i - \overline{x})^T = \frac{1}{n} \displaystyle\sum\limits_{i=1}^n {y'_i} {x'_i}^T
\end{equation}
                    
Using SVD one can decompose the correlation matrix $C = U W V^T$ where $U$ and $V$ are orthogonal matrices and $W$ is a diagonal matrix containing the singular values of C. By substituting the SVD of $C$ into Equation \ref{equ:trace}, we get: \par

\begin{equation}
tr (R^T C) = 
tr \{ R^T U W V^T\} = 
tr \{ V^T R^T U W\}
\end{equation}
            
We now define a new matrix $Q$ as: \par

\begin{equation} \label{equ:Q_def}
Q = V^T R^T U
\end{equation}

Thus we now have to maximise: \par
\begin{equation} \label{equ:tr(qw)}
 tr (R^T C) = tr(QW)
\end{equation}

Since $W$ is a diagonal matrix, the result of Equation \ref{equ:tr(qw)} is only influenced by the values along the main diagonal of $Q$. Thus maximising Equation \ref{equ:tr(qw)} becomes the problem of maximising the values on the main diagonal of $Q$. \par
            
As $V$, $R$ and $U$ are orthogonal the same must hold for $Q$. The Euclidean vector norm of the main diagonal of $Q$ must be equal or less than $1$. Therefore, in order to maximise Equation \ref{equ:tr(qw)}, $Q$ has to be the identity matrix. \par
            
Going back to Equation \ref{equ:Q_def} and substituting $Q$ with $I$: \par

\begin{equation}              
I = V^T R^T U \rightarrow R V = U \rightarrow R = U V^T
\end{equation}
                
This solution fails in certain cases when the determinant of $R$ becomes $-1$, which makes it a reflection not a rotation. This has been resolved by \textit{Kanatani} and \textit{Challis}. After computing the SVD of $C$, we can maximise $tr(R^T C)$ (Equation \ref{equ:trace}) if: \par 

\begin{equation}
 R = U 
\begin{bmatrix}
1 & 0 & 0\\
0 & 1 & 0\\
0 & 0 & det(UV^T)
\end{bmatrix}
 V^T
\end{equation}
                
The method will no longer fail in the cases where $det(UV^T)$ is $-1$ as $R$ will have a determinant of $+1$. \par

Two main methods are used to increase the frame rate. Firstly based on the positions of the markers in the previous frame, a region of interest is defined to reduce the number of pixels in the blob detection phase as seen in Fig. \ref{fig:tracking}(b).  \par

\begin{figure*}[h!]
\centering
\vspace {2pt}
\subcaptionbox{}
[0.46\textwidth]{\includegraphics[width=0.46\textwidth]{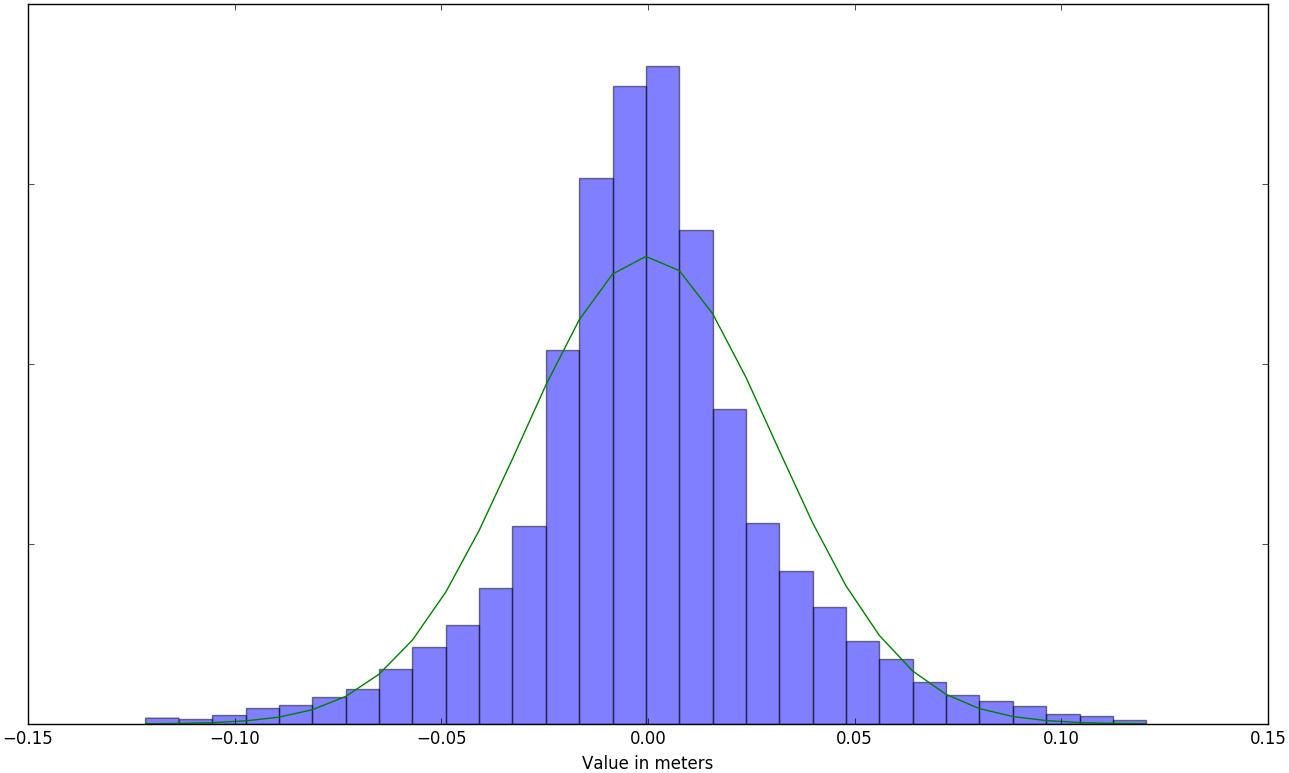}}
\subcaptionbox{}
[0.455\textwidth]{\includegraphics[width=0.455\textwidth]{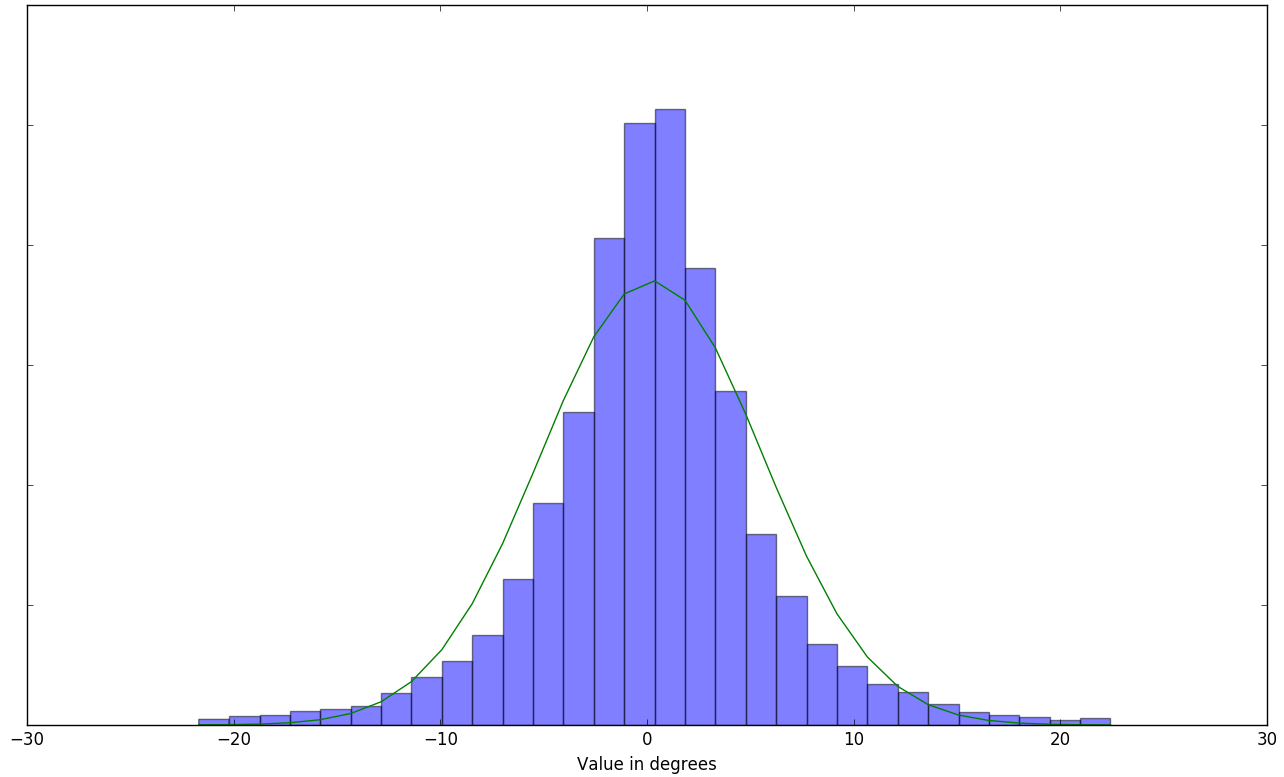}}
    \caption[]{(a) Distribution of the positional error of 10000 data points. The average absolute error is of 22.1mm (b) Distribution of the angular error of 10000 data points. The average absolute error is of $3.87^{\circ}$. In the dynamic case the error follows a Gaussian as per the central limit theorem due to multiple independent error sources.}
    \label{fig:dynamic_err}
\end{figure*}  

Secondly, we interpolate the position and orientation in between frames. We use simple linear interpolation of position assuming constant speed. For orientation, we use SLERP \cite{shoemake1985animating}, a method for linearly interpolating between quaternions. As the movement markers on the object are tied to the motion of the human hand, the system is not highly dynamic and thus such simple linear interpolations prove adequate. \par

The tracking runs in 46 fps in static conditions and 41 fps in dynamic conditions. The entire workflow split into programming jobs is visible in Fig. \ref{fig:workflow}. The computing job runs on two threads, while the prediction and streamer jobs each run on one.\par

\section{EXPERIMENTS AND RESULTS}

We performed two groups of tests. A static one where both the HoloLens and the tracked input device are static and a dynamic one where both the HoloLens and the input device move. To get directly the precision of the tracking we skipped the referencing step using the point cloud. We took 1000 static samples before each test to find a transformation that minimises the root mean square error between the ground truth position of the input device obtain via the ART-3 tracking system and the transform of the input device position in the HoloLens coordinate system. \par  

For measuring the precision in static conditions, the input device and the Hololens were both placed in stationary positions one arm-length away. After calibrating the world coordinate of the ART system, we gathered 5.000 samples of pose data for the pen pose seen from the Hololens and the ART tracker. \par
    
To measure the precision in dynamic conditions, we moved randomly through the room again and gathered 10.000 samples of pose data in 5 experiments of 2.000 samples each. The referencing was performed at the start as previously described. \par

The static tests have shown an average absolute positional error of 1.9 mm and an average absolute angular error of $0.37^{\circ}$. In the dynamic case an average absolute positional error of 22.1 mm and an average absolute angular error of $3.87^{\circ}$ were observed. Fitting a Gaussian distribution to the positional errors gave the median of the positional error of -0.1 mm and the standard deviation of 30.7 mm. The angular error distribution had a median of $0.22^{\circ}$ and a standard deviation of $5.39^{\circ}$. \par

\subsection{Discussion}

Interesting to see is the non-Gaussian distribution of the stationary tracking error. This is most likely due to the fact that the main error contribution is the IMU drift. Therefore the quality of the tracking was based purely on the average absolute error. On the other hand, while moving, independent error sources such as IMU drift, visual odometry errors etc. add to a Gaussian distribution as per the central limit theorem. Thus the error in the dynamic case does follow a Gaussian distribution. The biases found in the error distribution can then be used as ad-hoc corrections to the obtained tracking data. \par

The main contribution to the tracking error in the dynamic case seems to be the localisation error of the HoloLens, as is noticeable from the difference between the static and dynamic cases. In most cases the user will not move significantly while inputting trajectories, thus the error is expected to usually be much closer to the static case than the fully dynamic case. \par

Compared to four other state of the art commercial tracking devices (OptiTrack Flex3, Qualisys ARQUS and MIQUS and the Vicon Vantage), the static position error is four times higher than the average of the four devices (0.5mm) while the dynamic one is significantly worse at 45 times higher. The achieved frame rate however is only 2.5 times lower than the average of 100 frames per second. \par

In \cite{francois2011handprecision}, \textit{B{\'e}rard et al.} tested human input precision from various devices. It was shown that mouse and stylus type devices have a human input precision of around 0.5 mm while free-space devices have a precision of 5 mm. Thus the precision of the tracking in free space without significant HoloLens motion is adequate. Assuming that when the input device contacts a surface the precision will be approximately same to the mouse or stylus, the tracking precision will need to be improved in those cases. \par

There are ways to further reduce the tracking error to achieve the necessary precision. The most obvious one is to use newer generations of HMDs, like the HoloLens 2, which have improved tracking capability. Additionally the use of sensors on the robot itself may improve accuracy dramatically. In \cite{hartmann2019machining} it was shown that a laser line sensor on the robot can be used to vastly improve the accuracy of user input to achieve sub-millimetre precision. The input was given through the same input-device tracked here.\par

\section{CONCLUSION}

Here we presented an inside-out tracking system for IR-tracked input devices using the Microsoft HoloLens. The combination of AR and IR-tracked input devices has been a staple of intuitive robot programming research and has shown dramatic decreases in programming time and increase in precision even when non-experts used the system. Such systems, however, are expensive, static and inflexible, requiring extensive calibration before use. Our system would mitigate the disadvantages present by providing a mobile, robot-agnostic system that can both provide AR interaction and IR tracking capabilities. \par

Experiments have shown that the system performs adequately. In free-space, without any major HoloLens motion, the tracking precision is in the range of human motion precision. Human motion precision, however can increase up to 0.5mm when the input device is in contact with a surface. \par

Still future prospects are exciting. The system will be tested with non-experts in various programming tasks to see how and if the intuitiveness of programming improved with our system. Better model-based prediction and filtering will be researched to help improve tracking accuracy. Using newer HMDs such as the HoloLens 2 would decrease the main error contribution which is the HMD's localisation error. Our system can directly be deployed on the HoloLens 2 to test the influence of the better localisation. Finally, using various sensor mounted on the robot end-effector to improve the user input is a promising way to greatly improve accuracy as demonstrated by \textit{Hartmann et al.} \cite{hartmann2019machining}. \par   

\addtolength{\textheight}{-12cm}   




\section*{ACKNOWLEDGMENT}
This work was supported by the Federal Ministry of Education and Research of Germany within the Program Research for Civil Security Project ROBDEKON (Funding No. 13N14678).


\bibliographystyle{IEEEtran}
\bibliography{bib}

\begin{thebibliography}{10}
\providecommand{\url}[1]{#1}
\csname url@samestyle\endcsname
\providecommand{\newblock}{\relax}
\providecommand{\bibinfo}[2]{#2}
\providecommand{\BIBentrySTDinterwordspacing}{\spaceskip=0pt\relax}
\providecommand{\BIBentryALTinterwordstretchfactor}{4}
\providecommand{\BIBentryALTinterwordspacing}{\spaceskip=\fontdimen2\font plus
\BIBentryALTinterwordstretchfactor\fontdimen3\font minus
  \fontdimen4\font\relax}
\providecommand{\BIBforeignlanguage}[2]{{%
\expandafter\ifx\csname l@#1\endcsname\relax
\typeout{** WARNING: IEEEtran.bst: No hyphenation pattern has been}%
\typeout{** loaded for the language `#1'. Using the pattern for}%
\typeout{** the default language instead.}%
\else
\language=\csname l@#1\endcsname
\fi
#2}}
\providecommand{\BIBdecl}{\relax}
\BIBdecl

\bibitem{schraft2006need}
R.~D. Schraft and C.~Meyer, ``The need for an intuitive teaching method for
  small and medium enterprises,'' \emph{VDI BERICHTE}, vol. 1956, p.~95, 2006.

\bibitem{hein2009intuitive}
B.~{Hein} and H.~{Wörn}, ``Intuitive and model-based on-line programming of
  industrial robots: New input devices,'' in \emph{2009 IEEE/RSJ International
  Conference on Intelligent Robots and Systems}, 2009, pp. 3064--3069.

\bibitem{Zaeh2007Interactive}
M.~F. {Zaeh} and W.~{Vogl}, ``Interactive laser-projection for programming
  industrial robots,'' in \emph{2006 IEEE/ACM International Symposium on Mixed
  and Augmented Reality}, Oct 2006, pp. 125--128.

\bibitem{Reinhart2007projection-based}
G.~{Reinhart}, W.~{Vogl}, and I.~{Kresse}, ``A projection-based user interface
  for industrial robots,'' in \emph{2007 IEEE Symposium on Virtual
  Environments, Human-Computer Interfaces and Measurement Systems}, June 2007,
  pp. 67--71.

\bibitem{gaschler2014intuitive}
A.~Gaschler, M.~Springer, M.~Rickert, and A.~Knoll, ``Intuitive robot tasks
  with augmented reality and virtual obstacles,'' in \emph{Robotics and
  Automation (ICRA), 2014 IEEE International Conference on}.\hskip 1em plus
  0.5em minus 0.4em\relax IEEE, 2014, pp. 6026--6031.

\bibitem{REINHART2008programing}
\BIBentryALTinterwordspacing
G.~Reinhart, U.~Munzert, and W.~Vogl, ``A programming system for robot-based
  remote-laser-welding with conventional optics,'' \emph{CIRP Annals}, vol.~57,
  no.~1, pp. 37 -- 40, 2008. [Online]. Available:
  \url{http://www.sciencedirect.com/science/article/pii/S0007850608000917}
\BIBentrySTDinterwordspacing

\bibitem{Leutert2013}
\BIBentryALTinterwordspacing
F.~Leutert, C.~Herrmann, and K.~Schilling, ``A spatial augmented reality system
  for intuitive display of robotic data,'' in \emph{Proceedings of the 8th
  ACM/IEEE International Conference on Human-robot Interaction}, ser. HRI
  '13.\hskip 1em plus 0.5em minus 0.4em\relax Piscataway, NJ, USA: IEEE Press,
  2013, pp. 179--180. [Online]. Available:
  \url{http://dl.acm.org/citation.cfm?id=2447556.2447626}
\BIBentrySTDinterwordspacing

\bibitem{notheis2012ar-based}
S.~{Notheis}, W.~{August}, B.~{Hein}, and H.~{Woern}, ``Ar-based approach for
  evaluation of new model-based control algorithms,'' in \emph{ROBOTIK 2012;
  7th German Conference on Robotics}, May 2012, pp. 1--5.

\bibitem{Alexa2003MLS}
M.~{Alexa}, J.~{Behr}, D.~{Cohen-Or}, S.~{Fleishman}, D.~{Levin}, and C.~T.
  {Silva}, ``Computing and rendering point set surfaces,'' \emph{IEEE
  Transactions on Visualization and Computer Graphics}, vol.~9, no.~1, pp.
  3--15, Jan 2003.

\bibitem{puljiz2020hololens}
D.~{Puljiz}, F.~{Krebs}, F.~{B{\"o}sing}, and B.~{Hein}, ``What the hololens
  maps is your workspace: Fast mapping and set-up of robot cells via head
  mounted displays and augmented reality,'' in \emph{2020 IEEE/RSJ
  International Conference on Intelligent Robots and Systems (IROS)}, 2020.

\bibitem{TUW-210294}
M.~Mehling, ``Implementation of a low cost marker based infrared light optical
  tracking system,'' Master's thesis, Institute for Software Technology {\&}
  Interactive Systems, 2006.

\bibitem{article}
B.~Horn, H.~Hilden, and S.~Negahdaripour, ``Closed-form solution of absolute
  orientation using orthonormal matrices,'' \emph{Journal of the Optical
  Society of America A}, vol.~5, pp. 1127--1135, 07 1988.

\bibitem{4767965}
K.~S. {Arun}, T.~S. {Huang}, and S.~D. {Blostein}, ``Least-squares fitting of
  two 3-d point sets,'' \emph{IEEE Transactions on Pattern Analysis and Machine
  Intelligence}, vol. PAMI-9, no.~5, pp. 698--700, 1987.

\bibitem{Umeyama1991LeastSquaresEO}
S.~Umeyama, ``Least-squares estimation of transformation parameters between two
  point patterns,'' \emph{IEEE Trans. Pattern Anal. Mach. Intell.}, vol.~13,
  pp. 376--380, 1991.

\bibitem{291441}
K.~{Kanatani}, ``Analysis of 3-d rotation fitting,'' \emph{IEEE Transactions on
  Pattern Analysis and Machine Intelligence}, vol.~16, no.~5, pp. 543--549,
  1994.

\bibitem{c0ae9ca782784f238570e1fc0c232744}
J.~Challis, ``\BIBforeignlanguage{English (US)}{A procedure for determining
  rigid body transformation parameters},'' \emph{\BIBforeignlanguage{English
  (US)}{Journal of Biomechanics}}, vol.~28, no.~6, pp. 733--737, Jun. 1995.

\bibitem{shoemake1985animating}
K.~Shoemake, ``Animating rotation with quaternion curves,'' in
  \emph{Proceedings of the 12th annual conference on Computer graphics and
  interactive techniques}, 1985, pp. 245--254.

\bibitem{francois2011handprecision}
F.~B{\'e}rard, G.~Wang, and J.~R. Cooperstock, ``On the limits of the human
  motor control precision: The search for a device's human resolution,'' in
  \emph{Human-Computer Interaction -- INTERACT 2011}, P.~Campos, N.~Graham,
  J.~Jorge, N.~Nunes, P.~Palanque, and M.~Winckler, Eds.\hskip 1em plus 0.5em
  minus 0.4em\relax Berlin, Heidelberg: Springer Berlin Heidelberg, 2011, pp.
  107--122.

\bibitem{hartmann2019machining}
D.~{Hartmann}, M.~{Mende}, D.~{Štogl}, B.~{Hein}, and T.~{Kröger},
  ``Robot-based machining of unmodeled objects via feature detection in dense
  point clouds,'' in \emph{2019 IEEE/RSJ International Conference on
  Intelligent Robots and Systems (IROS)}, 2019, pp. 7777--7783.

\end{thebibliography}

\end{document}